\documentclass{article}

\usepackage{arxiv}
\usepackage[utf8]{inputenc}
\usepackage[T1]{fontenc}
\usepackage{hyperref}
\usepackage{url}
\usepackage{booktabs}
\usepackage{amsfonts}
\usepackage{nicefrac}
\usepackage{microtype}
\usepackage[dvipsnames]{xcolor}
\usepackage{graphicx}
\AtBeginDocument{\setkeys{Gin}{height=0.72\textheight,keepaspectratio}}
\usepackage{amsmath,amssymb}
\usepackage[ruled,vlined]{algorithm2e}
\usepackage{authblk}
\usepackage{float}

\title{SA-RAG: Leveraging Spreading Activation for Improved Document Retrieval in Knowledge-Graph-Based RAG Systems}
\author[2]{Jovan Pavlović}
\author[1,2,3]{Miklós Krész}
\author[1,2]{László Hajdu}
\affil[1]{\textbf{InnoRenew CoE, UP IAM, University of Primorska}}
\affil[2]{\textbf{UP FAMNIT, University of Primorska}}
\affil[3]{\textbf{University of Szeged, Department of Applied Informatics}}

\providecommand{\appendices}{\appendix}

\begin{document}
\maketitle
\begin{abstract}
Despite initial successes and a variety of architectures, retrieval-augmented generation (RAG) systems still struggle to reliably retrieve and connect the multi-step evidence required for complicated reasoning tasks. Most of the standard RAG frameworks regard all retrieved information as equally reliable, overlooking the varying credibility and interconnected nature of large textual corpora. GraphRAG approaches offer potential improvement to RAG systems by using knowledge graphs to capture entity relationships and enable multi-step reasoning, but their impact is often limited by the need for accurate graph representations (often costly to curate or unreliable to induce automatically), as well as by their continued dependence on large language models (LLMs) to navigate the graph and guide evidence selection.  In this paper, we propose the SA-RAG framework that uses a spreading activation algorithm to retrieve information from a corpus of documents connected by an automatically constructed heterogeneous knowledge graph. By operating on this heterogeneous structure, the method reduces reliance on semantic knowledge graphs that can be incomplete due to information loss during the graph construction process, while avoiding LLM-guided graph traversal. Experiments on multi-hop question-answering benchmarks show that the proposed approach achieves competitive performance comparable to state-of-the-art graph-based solutions.
\end{abstract}


\section{Introduction}
Initially proposed in~\cite{guu2020retrieval, lewis2020retrieval}, Retrieval-Augmented Generation (RAG) has become a popular technique for improving the capabilities of Large Language Models (LLMs) across various tasks, including question answering and code generation.  The basic idea of RAG is to provide LLMs with access to an external, easily updatable knowledge source by integrating it with the Information Retrieval (IR) component. This technique aims to improve the quality of LLM responses by reducing hallucinations and enhancing text creation accuracy and coherence. Numerous enhancements to the foundational RAG paradigm have been further proposed, ranging from innovative pipeline designs to improvements in individual components. Additionally, several surveys have been conducted to classify, decompose, and identify key elements within the myriad of proposed methodologies, helping track their evolution and highlighting promising future directions~\cite{zhao2024retrieval, fan2024survey, gao2023retrieval, gupta2024comprehensive}. 

A particularly active research area explores the combination of RAG with graph-based systems that represent diversified and relational information. The goal of these integrations is to improve performance on complex tasks involving large, structured, and interconnected knowledge corpora, resulting in methodologies sometimes referred to as GraphRAG. Comprehensive surveys covering the background, components, downstream tasks, and industrial use cases of GraphRAG, as well as the technologies and evaluation methods used, are available in~\cite{peng2024graph, han2024retrieval}. 

As mentioned in~\cite{peng2024graph}, knowledge graph question answering (KGQA) is a key natural language processing task that has motivated the development of knowledge-graph-based systems aimed at responding to user queries through structured reasoning over predefined knowledge graphs~\cite{zhang2021neural, lavrinovics2025knowledge, pan2024unifying}. The multi-hop question answering (MHQA) task is a related but broader concept, commonly applied in areas such as academic research, customer support, and financial or legal inquiries, where a comprehensive analysis of multiple textual documents is required. In this scenario, given a document corpus, the QA system generates responses to user queries by reasoning across multiple sources, involving sequential reasoning steps where responses at one step may rely on answers from prior steps. Several standard and graph-based RAG methodologies have been proposed to enhance the performance of LLMs on this specific task~\cite{zhang2023end, jeong2024adaptive, wang2024knowledge, jimenez2024hipporag}. Despite significant advancements, most of these approaches still rely primarily on standard or iterative retrieve-read-answer workflows, which are enhanced by optimized pipeline components, LLM decision-making, or LLM-guided graph traversal~\cite{singh2025agentic, fan2024survey}. However, hallucination and weak faithfulness remain unresolved problems in the context of MHQA. LLM-guided iterative retrieval may fail due to myopic knowledge exploration, retrieving context based on partial reasoning from previous steps, which can be incorrect or contain fabricated information. Similarly, one-step RAG systems with optimized retrieval components may fetch more useful evidence in the retrieved context, but they often introduce a lot of noise or fail to capture information from "bridge" documents whose entities or phrases are not mentioned in the input query. Thus, although these methods improve vanilla RAG, their underlying paradigm remains fundamentally the same, thereby overlooking powerful and well-established IR algorithms that can automatically identify relationships between documents and adapt retrieval for multi-hop query scenarios.

Motivated by this gap, our research goal is to develop a retrieval mechanism for RAG systems that can identify interconnected evidence needed for multi-step reasoning without relying on an LLM to control traversal over a knowledge graph. To this end, we propose SA-RAG, a RAG framework that integrates Spreading Activation (SA), a search method for associative networks originally developed in cognitive psychology. SA-RAG operates on a hybrid graph structure combining a knowledge graph with textual information from knowledge documents. It executes the SA algorithm through the knowledge graph to retrieve documents that contain complementary evidence for answering multi-hop questions. Our experiments suggest that SA-RAG performs on par with evaluated state-of-the-art approaches, achieving comparable or better results on standard MHQA benchmarks. 

To the best of our knowledge, only one prior paper has integrated RAG and SA-based methodologies~\cite{wu2025kg}. However, our approach differs in several key respects. First, the mentioned system relies on human-crafted knowledge graphs, whereas our pipeline includes an automated knowledge-graph construction phase. Second, it uses a prompted LLM to perform the SA procedure and expand the initially retrieved subgraph; by contrast, we perform SA automatically by exploring the knowledge graph in a breadth-first manner and spreading the activation based on edge weights assigned by an embedding model. Finally, the prior method fine-tunes the LLM to improve its instruction-following abilities, while our method requires no retraining. Overall, we believe our methodology provides an innovative direction for enhancing retrieval-augmented generation by leveraging a cognitively inspired mechanism that emphasizes associative relevance rather than surface-level similarity.

We summarize the key contributions of our work as follows:

\begin{enumerate}
    \item We introduce a knowledge-base indexing strategy that uses an LLM prompted to construct text-attributed knowledge graphs, linking documents and enabling the use of graph-traversal techniques during document retrieval.
    \item We introduce a novel integration of the Spreading Activation algorithm with LLMs, aiming to enhance grounded reasoning over large corpora of interconnected textual documents.
    \item We demonstrate notable performance improvements of our pipeline over standard RAG frameworks through experiments on two well-known multi-hop QA benchmarks. The results show that our SA-based document retrieval method enables more efficient exploration of the knowledge corpus than dense-vector retrieval alone, going beyond mere semantic similarity and enabling LLMs to produce more accurate answers to complex multi-hop questions.
    \item Given the fact that our experiments are performed with small open-weight models, which require less computational power, and yet we achieved significant results, we highlight the potential for deploying high-performance reasoning systems in resource-constrained environments.
\end{enumerate}

The rest of this paper is organized as follows. Section 2 reviews the background and related work. Section 3 presents the motivation behind the proposed SA-RAG method alongside its implementation details. Sections 4 and 5 describe the experimental setup and results, respectively. Finally, Section 6 concludes the paper by discussing limitations and future directions. 
The code used in our experiments is available at \url{https://github.com/jomibg/sa-rag-async}

\section{Background and Related Work}

This section reviews the main lines of research that the proposed method builds on. We begin by discussing how RAG has evolved to support multi-hop reasoning. We then outline how GraphRAG systems use graph structure to guide retrieval and generation. Finally, we introduce spreading activation and explain how its controlled variants can uncover "bridge" evidence that standard similarity searches miss.

\subsection{RAG and Multi-Hop Question Answering}
Although traditional RAG systems enhance LLM capabilities across both standard and knowledge-intensive NLP tasks, they encounter several critical limitations when handling complex reasoning. This is particularly evident in MHQA~\cite{mavi2024multi}, where the system is required to reason across multiple pieces of information or documents to answer complicated questions. Traditional systems rely on simple, single-shot retrieval, which often fails to capture all necessary information or may even overload the LLM’s context window with unnecessary or incorrect data. Numerous additions are proposed through the evolution of RAG paradigms with the aim of overcoming naive  "retrieve-then-generate" methodology, which, according to the survey by Fan et al., can be classified into two categories: \textit{training-free} and \textit{training-based} approaches. Training-free approaches aim to optimize RAG pipelines through techniques such as prompt engineering, in-context learning, improved organization of data in the knowledge corpus, or agentic design~\cite{singh2025agentic}. In contrast, training-based methods rely on fine-tuning the retriever and generator components to boost the system’s performance on downstream tasks.

Since our methodology does not require pre-training of any RAG components, we highlight a few training-free approaches that have been shown to enhance the performance of RAG systems in multi-hop question answering.

Multiple advanced RAG techniques use \textit{query expansion} as a means to improve retrieval steps by generating richer, semantically varied reformulations of the original question to match more relevant evidence from the knowledge corpus.

Articles~\cite{wang2023query2doc, gao2023precise} propose methods that first generate plausible answer-like passages for the input query and then embed them using sparse or dense encoders to match semantically similar real documents from the knowledge base.

Papers~\cite{ma2023query, fu2021decomposing} suggest that complex, implicit, or multi-hop queries should be rewritten, disambiguated, or decomposed before retrieval to optimize the fetched context.

Several papers have discussed the implementation of iterative retrieval procedures to enhance system performance on multi-hop questions, suggesting that retrieval should be viewed as an operation interleaved with the LLM’s own intermediate reasoning.

In~\cite{trivedi2022interleaving} authors propose a RAG system integrated with Chain‑of‑thought (CoT), a prompting technique, where each CoT step becomes the query for the next retrieval. Similarly, ~\cite{kim2023tree} proposes a system based on Tree‑of‑Thoughts (ToT) prompting framework; starting from an underspecified root question, the model generates candidate clarification subquestions, prunes uninformative branches, and, for each surviving node, runs a retrieval round targeted to that clarification, so that by the time the leaves are "answered", the system has accumulated disjoint, high-precision evidence for answering the original query.

Several papers propose integrating advanced RAG mechanisms with reflection modules that allow LLMs to determine the most suitable retrieval strategy based on the input query~\cite{jeong2024adaptive, wang2023self, asai2024self, crag}. By reflecting on the question, the LLM can choose whether to perform simple retrieval, engage in iterative retrieval steps, or rely on its own internal knowledge. Some systems can dynamically guide the direction and number of retrieval steps, evaluate the relevance of retrieved content~\cite{asai2024self}, or make use of external tools such as online search~\cite{crag}.

\subsection{GraphRAG}
The GraphRAG paradigm~\cite{peng2024graph, han2024retrieval} aims to extend RAG systems by leveraging the structured relationships and hierarchical organization of information in graph data to improve multi-hop reasoning and contextual understanding.

One of the first successful GraphRAG systems, introduced in~\cite{edge2024local}, was designed for Query-Focused Summarization (QFS), where, provided a query and a document corpus, the system seeks to compress the contents of multiple documents, gathering the identified themes and providing a unified summary of the entire document corpus. The proposed approach focuses on the connection between the summarization task and the inherent modularity of graphs, which allows partitioning of graphs into communities. LLM is first instructed to create a knowledge graph from source texts by extracting entities and relationships between them. The system then uses the Leiden community detection algorithm to partition this graph into hierarchical communities. Finally, each community is summarized separately, and the summaries are combined to produce a final, comprehensive response to the user's question.

With respect to MHQA, few works have received significant attention for effectively integrating graph-based retrieval and reasoning within the RAG. 

LightRAG proposed in~\cite{guo2024lightrag} firstly builds a lightweight knowledge graph over a document corpus by using an LLM to extract entities and relations from chunks. For each node and edge, it then constructs a key–value entry, where the key is a short textual label and the value is a concise summary derived from the source documents. At query time, another language model analyses the input and produces two sets of query terms, one focusing on concrete objects and relations, and another capturing broader topics. Terms from the first set are embedded and used for vector search over keys tied to entity nodes, while the broader theme terms are used to match keys associated with relation edges. The retrieved graph elements are expanded by one hop to include their neighbors, then the corresponding value texts are combined into a compact context that is passed to a generative model to produce the final answer.

In~\cite{g-retriever}, the authors present G-Retriever, a framework that integrates graph neural networks (GNNs) and LLMs. To retrieve information relevant for the input query, the system uses the Prize-Collecting Steiner Tree algorithm, which selects a connected subgraph by balancing the relevance of graph elements to the query against the cost of including additional edges. The system then efficiently responds to user queries by translating graph data to a textual representation and using GNN to transform its structural information into a soft prompt for the LMM. The graph encoder and projection layer, which are in charge of aligning subgraph information with the language model's hidden space, are fine-tuned via backpropagation based on the difference between the generated answer and the ground truth. This technique is evaluated on different tasks, ranging from commonsense reasoning and scene comprehension to knowledge-based question answering.

These systems differ, however, from our approach in that they search a constructed knowledge graph to gather relational knowledge, then convert this context into textual form before passing it to the LLM. In contrast, our methodology uses knowledge graph search to match textual documents directly without requiring an intermediate step of transforming graph data into text. 

Another MHQA pipeline is presented in~\cite{jimenez2024hipporag} and~\cite{gutierrez2025rag}. Their proposed approach, HippoRAG, creates a schemaless knowledge graph with instruction-tuned LLM. When a query is received, the language model extracts the key named entities from the query, which are then mapped to their associated nodes in the knowledge graph using similarity scores calculated by the retrieval encoder. These nodes serve as seeds for a graph search using the Personalized PageRank algorithm, which propagates activation through the network to identify nodes that may be associated with the provided question. Since this approach is similar to our solution in its use of a knowledge graph to retrieve relevant textual passages, we adopt it as a natural baseline in our experiments.

\subsection{Spreading Activation}

The study in ~\cite{quillian1967word} proposed that human semantic memory is organized as a network and was one of the first to implement a semantic network model of knowledge in a computer. In this framework, concepts and their properties were modeled as nodes in a network connected by links encoding semantic relationships, arranged hierarchically so that more specific concepts point to more general ones. Activation spreading was introduced as a mechanism to search this network in order to verify factual sentences (e.g., canary is a bird). This was achieved by activating a concept and letting that activation move through the network until it intersected with activation from another concept. Later in \cite{collins1969retrieval} the computational model was tested against human performance in sentence–verification tasks, where people’s response times to statements at different hierarchical levels (e.g. “a canary is a bird” vs. “a canary is an animal”) were compared to the model’s predictions, showing that longer network traversals correlate with slower verification times. This model was then modified in~\cite{collins1975spreading, anderson1983spreading} to account for varying association strengths and decay of activation over distance, dropping the requirement that the network is strictly hierarchical. The spreading mechanism was generalized so that activation spreads from a currently active concept to all linked concepts, in parallel, with strength depending on the link weight and fading with distance.

After its development in the field of cognitive psychology, SA was adapted in later works as a computational tool for information retrieval and recommender systems. In this context, original principles were repurposed so that input queries or user profiles trigger activation in a graph of terms, documents, or items, and the resulting activation pattern is used to rank candidates by their inferred relevance or interest.

In~\cite{salton1988use}, the authors argued that SA can be implemented in IR systems to overcome the limitations of standard TF-IDF-based approaches, particularly when the user’s wording does not match the wording of the documents and when relevant documents are only indirectly connected to the query. They proposed a system using network representation of the document corpus, where both documents and index terms are modeled as nodes, interconnected by various types of links: term–document links modeling the presence of a term in a document, term–term links connecting terms that frequently co-occur or are semantically related according to a thesaurus, and document–document links connecting documents that share co-occurring terms. In this setting, the SA procedure starts by taking the query terms or documents retrieved by the normal vector-space search and assigning them an initial activation value. The activation then propagates from every active node to its neighbors, with the amount passed along each link scaled by the link’s strength, and nodes whose accumulated activation exceeds a specified threshold are finally retrieved. However, the authors implemented and evaluated only a simplified version of the described procedure due to the computational costs of a full implementation at that time. They concluded that the spreading activation procedure is a useful complementary module to standard IR systems, but it needs to be constrained by limiting the number of steps to just a few hops from the original query nodes and by using a decay factor smaller than one. Later works expanded criticism of unconstrained SA along the same lines; in~\cite{berthold2009pure}, the authors showed that the pure SA algorithm, where activation propagates over all outgoing links with no limits on hops, fan-out, or thresholds, converges to a fixed state and retrieval results become nearly independent of the input query. This underscores that heuristic constraints in applied SA-based systems are not just pragmatic features but essential for the retrieval of query‐dependent results.

More recent applications in IR and recommendation systems, therefore, employ adapted, task-driven variants of SA, where the spread is guided by the user’s information need or context and then combined with conventional ranking steps. In~\cite{ngo2014discovering}, the authors propose a semantic text search method based on SA, which, given an input query, performs named entity recognition and relationship extraction, then maps the extracted elements to an ontology to represent the query as a concise structured pattern. The method then spreads activation only to concepts that match the relationships actually expressed in the query, as well as to closely related information such as alternative names or parent categories. Activated concepts are retrieved and appended to the original query to form an expanded query, which is then executed by a classical sparse-vector search engine, allowing ordinary document ranking to exploit both the user’s words and the newly discovered latent concepts.

Article~\cite{blanco2010adapting} describes a semantic, knowledge-based recommender built around an ontology that stores domain concepts, user interests, and item descriptors, and then runs a controlled spreading of activation from the user profile toward semantically adjacent items so that recommendations reflect both explicit preferences and latent relations present in the knowledge base. The system routes the activation only through specific, typed links in the ontology, assigning a weight to each link based on its type and finally ranking associated items according to their activation values and standard recommendation filters. Similarly, \cite{papneja2018context} extends ontology-based SA with contextual cues, such as current task or situational attributes, so that activation is fired not only from the long-term user profile but also from context nodes. Activation propagates only along meaningful ontology links, with decay and link weights, thereby avoiding uncontrolled diffusion through the network.

Beyond information retrieval~\cite{crestani1997application} and recommendation systems, SA has been successfully applied in several other domains, including natural language processing~\cite{pollack1982natural, tsatsaronis2007word}, robotics, and reinforcement learning~\cite{kono2019activation}.

\section{Methodology}
In this section, we formalize the retrieval task and objective. We begin with a motivational example in which we explain the main advantages of using spreading activation for semantic search on knowledge graphs in the presence of multi-hop questions. We then walk through the complete SA-RAG pipeline. Each component is explained in turn, from indexing and subgraph construction to the exact details of the information retrieval method, and finally, the answer generation. We also cover key parameters and design choices that shape the behavior of the system.

\subsection{Motivation}
\label{sub:motiv}
\begin{figure}[ht]
  \centering
  \includegraphics[width=\linewidth]{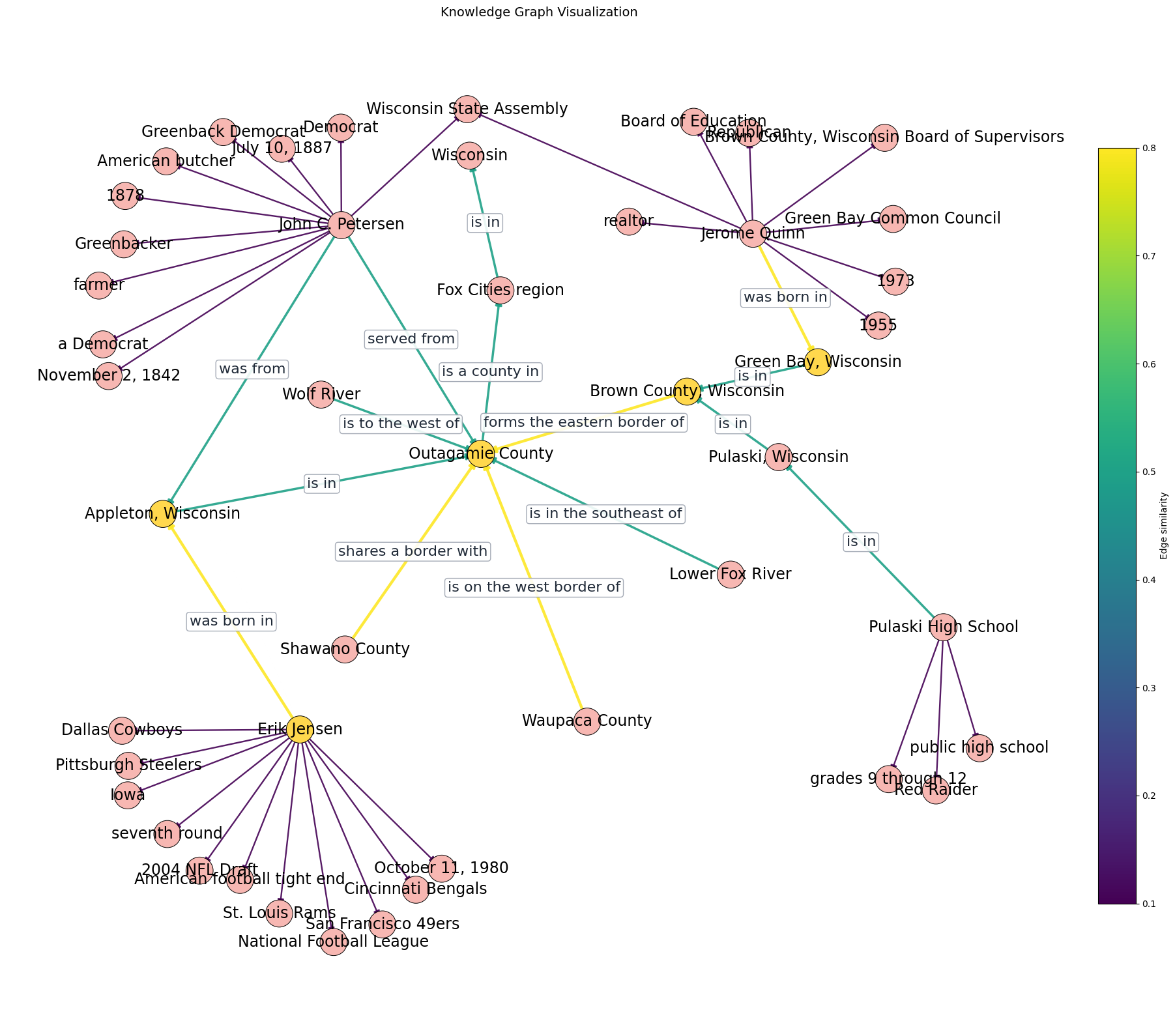}
  \caption{A knowledge graph representing the system's knowledge base for the example query. Edge weights reflect the semantic similarity between the input question and the relationships encoded by each edge. Yellow nodes represent \textit{golden} entities holding information necessary for answering the query.}
  \label{fig:kg_cs}
\end{figure}

Consider an IR system posed with the question \textbf{"What is the capital of the county that shares a border with the county that contains the birthplace of Erik Jensen?"} (query \textit{4hop1\_\_707078\_378185\_282674\_759393} from the MuSiQuE dataset). Assume the system's knowledge base can be represented as a weighted knowledge graph displayed in Figure~\ref{fig:kg_cs}, where the weights on edges represent semantic similarity between the query and the relationships that these edges encode, and imagine that each entity is associated with a textual description. In order to provide the context necessary for answering the question, the system must gather information about five entities lying along the shortest path from the node \textit{"Erik Jensen"} to the node \textit{Green Bay, Wisconsin} (the so-called \textit{golden} entities). A straightforward solution would be to select all the entities and reason through all of the textual descriptions; however, this would introduce a lot of irrelevant information into the deduction procedure. Moreover, if the working memory of the reasoning agent (which is the context window in the case of LLMs) is limited, it would be necessary to filter out some of the nodes.

Now suppose that we can collect at most 10 entities and the text search algorithm identifies the descriptions of nodes \textit{"Erik Jensen"} and \textit{"Outagamie County"} as the semantically most similar to the input question. Also, let the edge weights $w_e$ be calculated by the following simple rule:
\begin{itemize}
    \item If the corresponding relationship expresses location and contains the word "border" or "born", let $w_e := 0.8$
    \item If the relationship expresses location (such as "is in"), let $w_e := 0.5$
    \item Otherwise, let $w_e:=0.1$
\end{itemize}

Clearly, more nuanced edge weights can be derived by comparing latent representations of questions and entities, produced by state-of-the-art embedding models, but this simplified scheme is sufficient to illustrate the core idea. 

Now, if we want to collect the rest of the relevant entities, we would intuitively look for entities \textit{"close"} to the initial ones, or nodes that act as bridges \textit{"between"} the known pieces of information. Equivalent intuition is captured by the concept of node centrality in network science, which are numerical values or rankings meant to summarize a node's position and topological role within the complex network.

Different definitions of centralities have been proposed in the literature~\cite{centrality_zoo}, each designed to capture a different aspect of a node's importance, and several existing GraphRAG frameworks have incorporated such measures into their retrieval pipelines. Notably, the authors of HippoRAG~\cite{jimenez2024hipporag} proposed the use of Personalized PageRank (PPR)~\cite{page1998pagerank} score as a robust metric summarizing the relevance of nodes in the knowledge graph based on specific clues in the input query. Closeness centrality, defined as the reciprocal of the sum of the shortest path distances from a given node to a set of other nodes, provides a complementary, distance-based perspective on node importance. However, we argue that the SA algorithm subsumes both of these perspectives, providing relevance scores summarizing the intuitive notions of \textit{"closeness"} and \textit{"betweenness"}.

\begin{figure}[ht]
  \centering
  \includegraphics[width=\columnwidth]{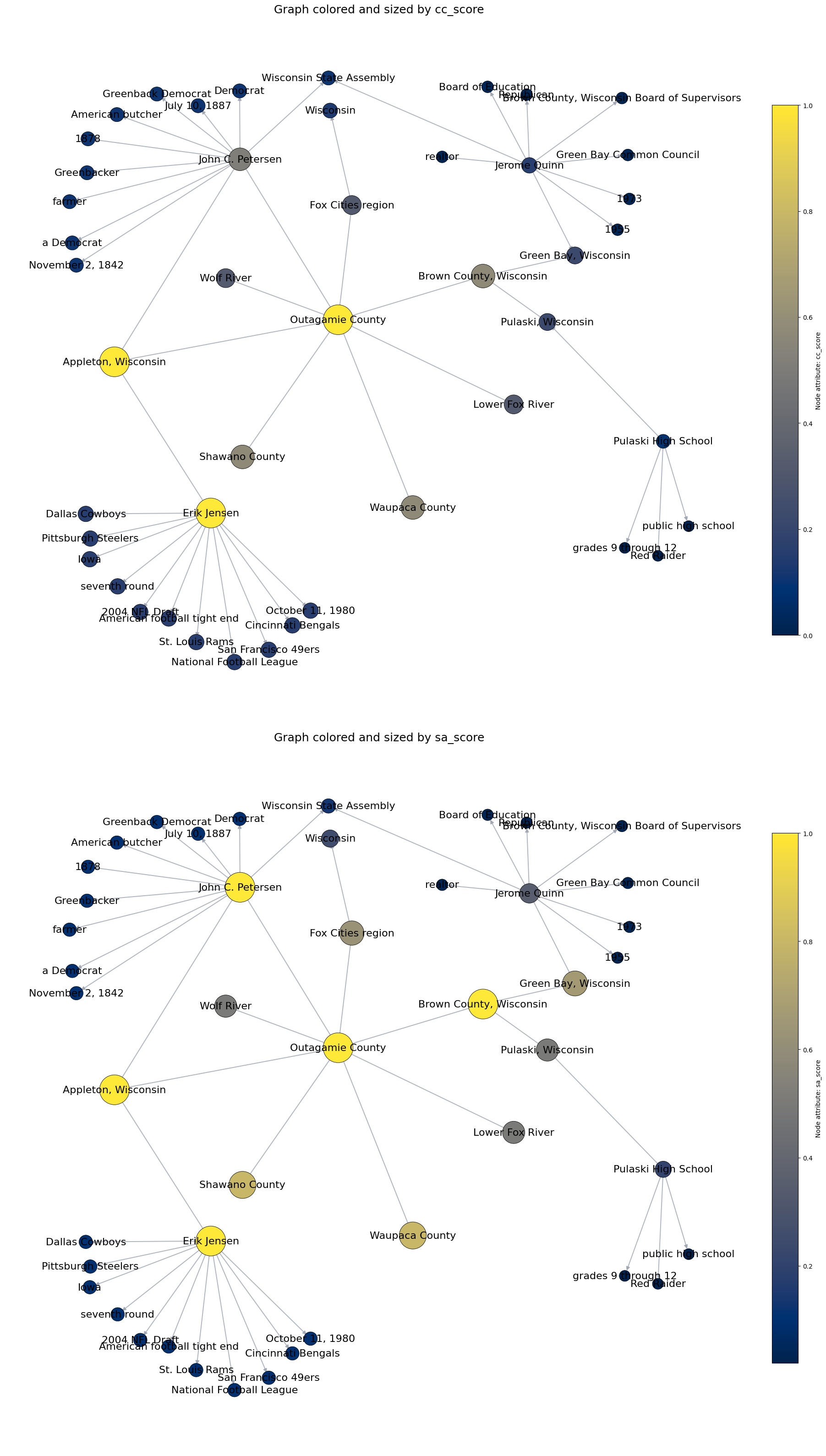}
  \caption{Visual comparison of (normalized) relevance scores assigned to knowledge graph nodes by closeness centrality (top) and the SA algorithm (bottom). Node size and color intensity both scale with the relevance score assigned by each method, larger and brighter nodes are ranked more relevant.}
  \label{fig:scores}
\end{figure}

Table~\ref{tab:entity_scores} displays node rankings and associated relevance scores calculated based on PPR, closeness centrality, and the SA algorithm detailed in Algorithm~1 (see Appendix \ref{app:motivational} for metric definitions). Because of the incomparable scale of metrics, we used min-max normalization to put closeness centrality and PPR scores into the interval from 0 to 1 (in order to compare scores to SA in the most intuitive way). As can be seen, in our example only SA collects all of the \textit{golden} entities associated with our input query. Similarly, Figure~\ref{fig:scores} shows our knowledge graph where the size and color of nodes indicate the relevance scores assigned by closeness centrality and SA, respectively. It is worth noting, however, that SA exhibits a tendency to assign high relevance scores to non-golden entities, raising the risk of false positives. As elaborated later in this section, we address this limitation through a set of additional mechanisms and hyperparameter choices, seeking to reduce the mentioned negative effects. (see Section \ref{sub:3f} and Appendix \ref{app:normalization}).

\begin{table}[ht]
\caption{Comparison of entities based on Closeness centrality, PPR score, and SA score.}
\label{tab:entity_scores}
\centering
\scriptsize
\setlength{\tabcolsep}{3pt}
\renewcommand{\arraystretch}{0.9}
\textbf{(a) Closeness centrality}

\begin{tabular}{@{}lr@{}}
\toprule
\textbf{Entity} & \textbf{Score} \\
\midrule
\textbf{Erik Jensen} & 1.0000 \\
\textbf{Outagamie County} & 1.0000 \\
\textbf{Appleton, Wisconsin} & 1.0000 \\
Shawano County & 0.5758 \\
Waupaca County & 0.5758 \\
\textbf{Brown County, Wisconsin} & 0.5758 \\
John C. Petersen & 0.5139 \\
Wolf River & 0.3137 \\
Lower Fox River & 0.3137 \\
Fox Cities region & 0.3137 \\
\bottomrule
\end{tabular}

\vspace{0.8em}

\textbf{(b) PPR score}

\begin{tabular}{@{}lr@{}}
\toprule
\textbf{Entity} & \textbf{Score} \\
\midrule
\textbf{Outagamie County} & 1.0000 \\
\textbf{Erik Jensen} & 0.8381 \\
\textbf{Appleton, Wisconsin} & 0.4479 \\
John C. Petersen & 0.2935 \\
\textbf{Brown County, Wisconsin} & 0.1874 \\
Shawano County & 0.1358 \\
Waupaca County & 0.1358 \\
Fox Cities region & 0.1328 \\
Wolf River & 0.0835 \\
Lower Fox River & 0.0835\\
\bottomrule
\end{tabular}

\vspace{0.8em}

\textbf{(c) SA score}

\begin{tabular}{@{}lr@{}}
\toprule
\textbf{Entity} & \textbf{Score} \\
\midrule
\textbf{Erik Jensen} & 1.0000 \\
\textbf{Outagamie County} & 1.0000 \\
\textbf{Brown County, Wisconsin} & 1.0000 \\
\textbf{Appleton, Wisconsin} & 1.0000 \\
John C. Petersen & 1.0000 \\
Shawano County & 0.8000\\
Waupaca County & 0.8000 \\
\textbf{Green Bay, Wisconsin} & 0.6640 \\
Fox Cities region & 0.6250 \\
Pulaski, Wisconsin & 0.5000 \\ 
\bottomrule
\end{tabular}

\end{table}

\subsection{Problem definition}
Given a question $q$ and a knowledge base $KB = \{D_1, \ldots, D_n\}$ containing a corpus of textual documents organized in some manner, the system generates an answer $a$ by retrieving a subset of relevant documents $\mathcal{D} \subseteq KB$. Formally, this can be expressed as:  
\begin{align*}
    \mathcal{D} &= R(q, KB) \\
    a &= \text{LLM}(q, \mathcal{D})
\end{align*}
where $R$ denotes the retrieval component of the system, and \textit{LLM} refers to the large language model responsible for generating the answer.  
The objective is to maximize the probability of producing the correct answer $a^*$, that is,  
\[
p(a^* = a \mid q, KB) = \sum_{\mathcal{D} \subseteq KB} p(a^* = a \mid q, \mathcal{D}) \, p(\mathcal{D} \mid q, KB).
\]
Since the generation process is not directly optimized, the task reduces to identifying the subset of documents $\mathcal{D}^*$ that maximizes the probability of providing the correct answer:
\[
\mathcal{D}^* = \arg\max_{\mathcal{D} \subseteq KB} p(a^* \mid q, \mathcal{D}) \, p(\mathcal{D} \mid q, KB).
\]

\subsection{System description}
We summarize the architecture of our RAG system enhanced with a spreading activation–based retrieval mechanism.  It consists of four main stages: indexing, subgraph fetching, spreading activation and document retrieval, and finally, answer generation. These stages are summarized below:
\begin{enumerate}
    \item \textit{Indexing}: In this phase, we break knowledge documents into chunks, create vector embeddings for these text chunks, and use them to build a knowledge graph of the entities, entity descriptions, and relationships mentioned. We also record references between each chunk and every entity it contains by storing the chunks as nodes in the graph, with outgoing links connecting them to the corresponding entity nodes.
    \item \textit{Subgraph fetching}: Given a query, we fetch a subgraph consisting of the most relevant nodes and relations based on the cosine similarity calculated between the vector embedding of the query and the embeddings of entity descriptions. Here, we also determine the set of initially activated entities according to their relevance as indicated by cosine similarity.
    \item \textit{Spreading activation and document retrieval}: We run spreading activation from the initially activated entities to obtain a subset of relevant nodes and fetch the textual paragraphs that mention these nodes. We also record the most relevant relations and combine their textual descriptions with the fetched documents.
    \item \textit{Answer generation}: The LLM then generates a grounded answer using the information from the fetched context.
\end{enumerate}

\begin{figure}[ht]
  \centering
  \includegraphics[width=\columnwidth]{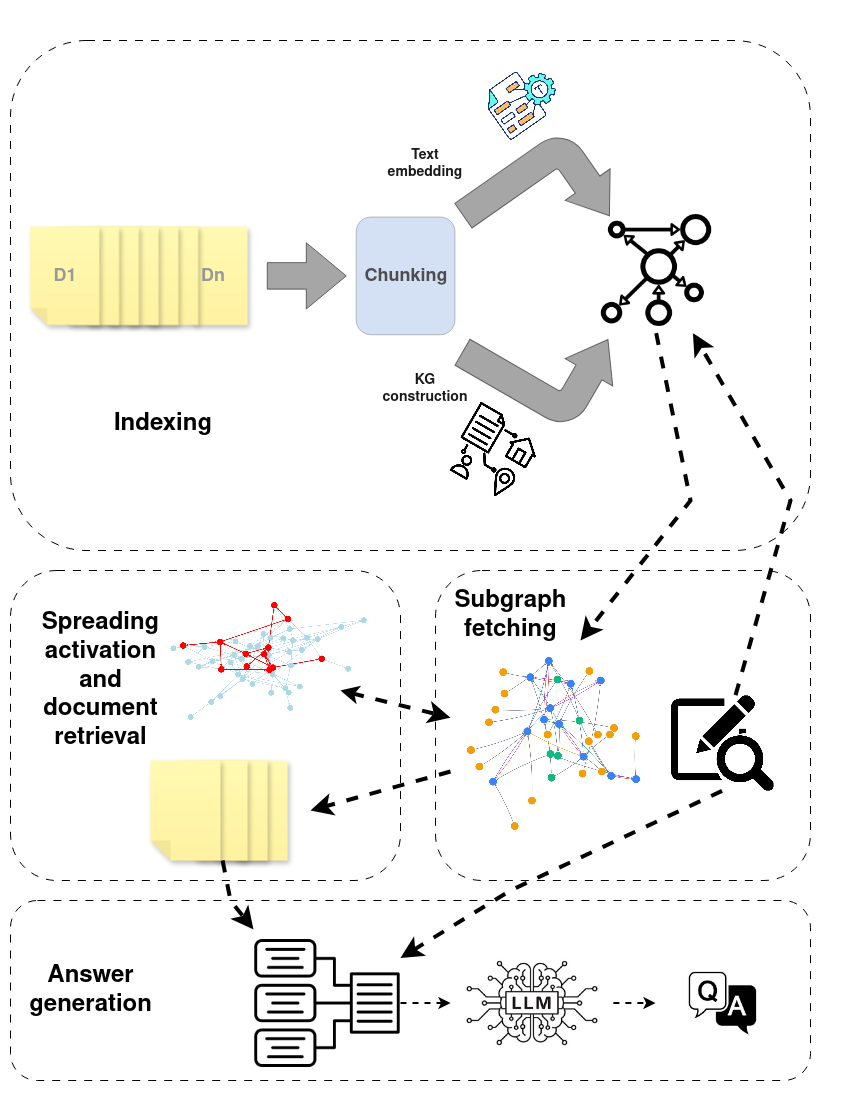}
  \caption{High-level overview of methodology}
  \label{fig:fig1}
\end{figure}

Figure~\ref{fig:fig1} provides a high-level overview of the proposed methodology.

\subsection{Indexing}
\begin{figure}[ht]
  \centering
  \includegraphics[width=\columnwidth]{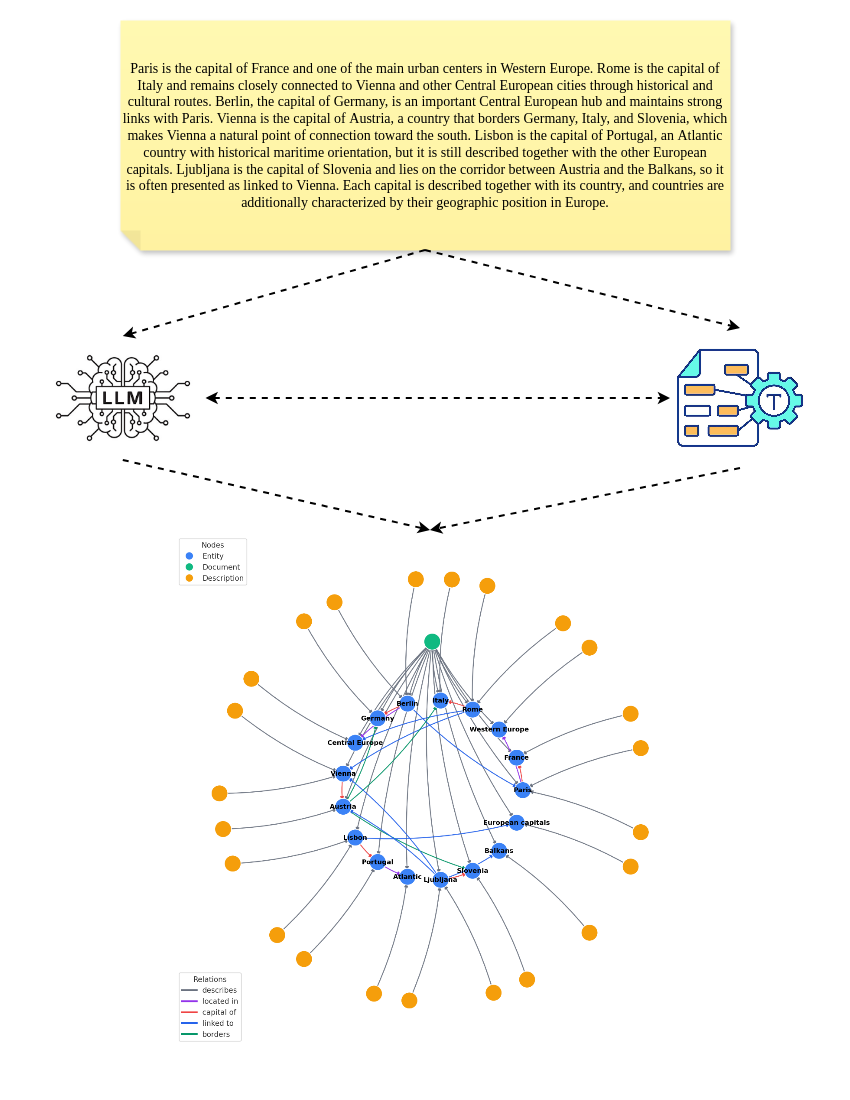}
  \caption{Knowledge graph creation during the indexing phase. The figure shows an example graph constructed from a single document describing the capital cities of European states. The document node is marked in green, entity description nodes are colored orange, and entity nodes are shown in blue.}
  \label{fig:fig2}
\end{figure}

We begin the indexing phase by performing word-based chunking by splitting input documents into chunks of 500 words with an overlap of 200 words between consecutive chunks. Each chunk is passed to the embedding model and then to a prompted LLM, which is instructed to extract entities, relations, and entity descriptions from the text. We construct a knowledge graph containing three types of nodes: entities, entity descriptions, and documents  (which represent the textual chunks). The graph also includes two types of links: \textit{describes} and \textit{related\_to}. The \textit{describes} links connect document and entity description nodes to their corresponding entities, while the \textit{related\_to} links connect two entities that are described as related within the text chunk. Additionally, each \textit{related\_to} link stores a string attribute holding the textual description of the relationship between the endpoint entities. Each entity node stores its \textit{name} as well as a list of alternative names within an \textit{aliases} attribute. To prevent duplicate entities, we check whether the \textit{name} or any \textit{aliases} of each newly extracted entity match those of an entity already in the knowledge graph. If the entity already exists, we create new entity description nodes and a \textit{describes} link connecting them to the corresponding entity. Additionally, we create vector embeddings for both entity description nodes and \textit{related\_to} links using the same embedding model applied to the textual chunks. The indexing process is illustrated in Figure~\ref{fig:fig2}.

\subsection{Subgraph fetching}
\begin{figure}
  \centering
  \includegraphics[width=\columnwidth]{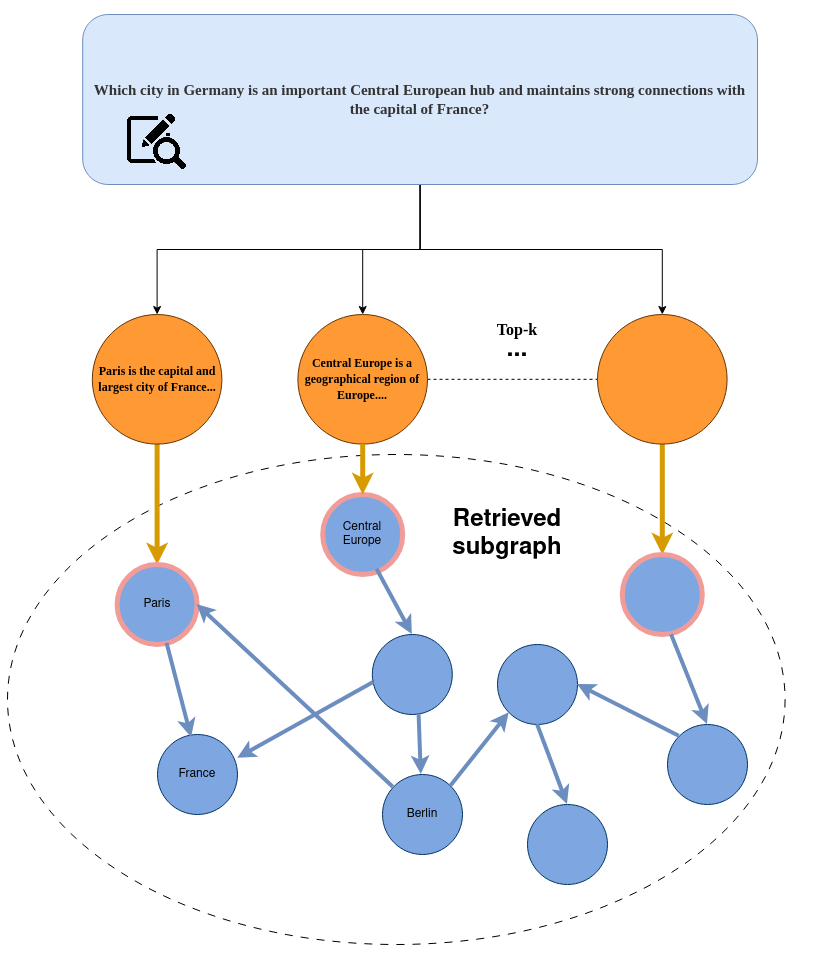}
  \caption{Subgraph fetching step: Orange nodes represent the \textit{top-k} entity description nodes, while entity nodes are shown in blue. Links of the \textit{describes} type are colored orange, whereas blue links represent \textit{related\_to} relations connecting entity nodes. Initially activated “seed” entities are indicated by a red border.}
  \label{fig:fig3}
\end{figure}

At this stage, we retrieve the subgraph consisting of entities and relations that are potentially relevant to answering the input query. Specifically, we first identify the top-$k$ entity descriptions whose embeddings have the highest cosine similarity to the query embedding. We then include the entities referenced by these descriptions through incoming \textit{describes} links and expand the graph by retrieving all entities and \textit{related\_to} links within their $n$-hop neighborhood. The “seed” entities matched in the first step are marked as initially activated for the spreading activation process, and each link is assigned a weight equal to the cosine similarity between the query and the link embedding. The process is illustrated in Figure~\ref{fig:fig3}.

\subsection{Spreading activation and document retrieval}
\label{sub:3f}
This stage simulates spreading activation on the retrieved subgraph, starting from the initially activated nodes identified in the previous step. Before the propagation begins, we rescale the edge weights by a linear factor $c$, using the formula $w' = max(0,\frac{w-c}{1-c})$. This adjustment helps prevent overactivation, which could otherwise result in a high number of activated yet irrelevant nodes, lowering system specificity and potentially leading to the context explosion (see Appendix \ref{app:normalization}).
The SA process operates as follows: initially, activated nodes are assigned an activation value of $a_i = 1$, while all other nodes get a value of $a_j = 0$. Activation then propagates outward from these initially activated nodes to their neighboring nodes in a breadth-first search manner, through the weighted links $w_{ij}$. In each iteration, the activation value of each target node is updated as $a_j = min(a_j + \sum_{i \in N(j)}a_i\cdot w_{ij},1)$, and after the process is complete, nodes whose activation exceeds the threshold $\tau_a$ are marked as activated. The entire procedure is summarized in Algorithm~\ref{alg:diffusion_process}.

\begin{algorithm}[h]
\caption{Spreading Activation algorithm on entity subgraph}
\label{alg:diffusion_process}
\DontPrintSemicolon

\KwIn{
  Adjacency dictionary $adj\_dict$ mapping entities to outgoing weighted arcs\\
  Set of initially activated entities $I_a$;\\
  Activation threshold $\tau_{\text{a}}$;
}
\KwOut{
  Set of activated entities $R_a$.
}

\BlankLine

\ForEach{$e \in \text{keys}(adj\_dict)$}{
    $activation\_value[e] \gets 0$\;
}

$Q \gets$ empty queue\;
\ForEach{$e \in I_a$}{
    $activation\_value[e] \gets 1$\;
    $Q.\text{append}(e)$\;
}
$visited \gets \emptyset$\;

\While{$Q$ is not empty}{
    $node \gets Q.\text{pop\_left}()$\;
    \If{$node \in visited$}{
        \textbf{continue}\;
    }
    $visited \gets visited \cup \{node\}$\;

    \ForEach{$(target, weight) \in adj\_dict[node]$}{
        $activation\_value[target] \gets \min\bigl(1,\; activation\_value[target] + weight\cdot activation\_value[node]\bigr)$\;
        \If{$target \notin visited$}{
            $Q.\text{append}(target)$\;
        }
    }
}

$R_a \gets \{\, e \mid activation\_value[e] > \tau_{\text{a}} \,\}$\;

\Return $R_a$\;
\end{algorithm}

After obtaining the set of activated entities, we fetch the set of documents $\mathcal{D} = \{D_{1},...D_{k_1}\}$  from the knowledge base, consisting of document nodes that have outgoing \textit{describes} links connecting them to the activated entity nodes. Here, we also perform filtering by removing documents whose cosine similarity to the query is lower than the pruning threshold  $\tau_{d}$. Additionally, we augment the set of retrieved documents with the textual representations of relevant relationships $r_1,... r_{k_2}$ between activated entities, as encoded in the links between them. In this way, we obtain the final context $\mathcal{C} = \{D_{1},...D_{k_1},r_1,... r_{k_2}\}$, which is then passed to the LLM in the next stage.

\subsection{Answer generation}
Finally, provided with the fetched context and the multi-hop query, the LLM is instructed to reason through the knowledge text in a step-by-step manner, verifying all arithmetic calculations if the question requires quantitative reasoning. It then outputs its reasoning and the final answer if it is able to deduce one based solely on the provided knowledge, or otherwise notes that the question is not answerable based on the retrieved information.

\subsection{Hyperparameter tuning}

We optimized main retrieval parameters of our GraphRAG pipeline using Optuna. For each benchmark dataset, parameters were tuned on a sample of 100 questions kept disjoint from the validation sets, with the knowledge graph constructed once from the corresponding corpus subset. For each hyperparameter trial, the system answered the questions using contexts retrieved under the trial’s parameter configuration. We fixed $\tau_a = 0.5$ as a neutral baseline; since $\tau_a$ interacts with the normalization factor $c$, fixing it avoids coupled search dimensions and isolates the impact of the remaining parameters.
\newpage
The optimized parameters were:
\begin{itemize}
    \item $k$: the number of entity descriptions used to activate the graph;
    \item $n$: the number of hops used to define the retrieved subgraph;
    \item $c$: a linear factor used to rescale edge weights;
    \item $\tau_d$: the document-pruning threshold;
\end{itemize}

The parameter configurations yielding the best performance are summarized in Table~\ref{tab:hyperparameters}.

\begin{table}[htbp]
  \centering
  \caption{Hyperparameters, search spaces, and optimal values.}
  \label{tab:hyperparameters}
  \begin{tabular}{lcccc}
    \toprule
    \textbf{Par.} & \textbf{Type} & \textbf{Search Space} & \textbf{MuSiQuE} & \textbf{2WikiMH}\\
    \midrule
    $k$ & Discrete & $[2, 10]$ & $6$ & $9$ \\
    $n$ & Discrete & $[2, 5]$ & $3$ & $3$ \\
    $c$ & Continuous & $[0.1, 0.9]$ & $0.45$ & $0.7$ \\
    $\tau_d$ & Continuous & $[0.1, 0.5]$ & $0.3$ & $0.25$ \\
    \bottomrule
  \end{tabular}
\end{table}

\section{Experiment Design and Evaluations}
This section describes how the proposed method is evaluated and what it is compared against. It first introduces the benchmarks and the evaluation setup, and then we proceed by defining the baseline RAG and GraphRAG methods used for comparison. 

\subsection{Benchmarks}
We evaluate our system on two well-known MHQA benchmarks: MuSiQue~\cite{musique} and 2WikiMultiHopQA~\cite{2wiki}.

One of the earliest and most widely used multi-hop QA datasets was HotpotQA~\cite{hp}, created through crowdsourced multi-hop questions over Wikipedia articles. However, it has faced considerable criticism for not consistently requiring genuine multi-hop reasoning, as many questions could be answered through dataset artifacts or inference shortcuts~\cite{hp_criticism1, hp_criticism2, hp_criticsm3}.

In response, 2WikiMultiHopQA~\cite{2wiki} refined the evaluation paradigm by combining unstructured Wikipedia text with structured Wikidata knowledge, thereby mitigating some of these issues. MuSiQue~\cite{musique} advanced this approach further by composing multi-hop questions in a bottom-up manner from a large pool of single-hop questions while enforcing strict connected reasoning. Therefore, in our experiments, we used 2WikiMultiHopQA and MuSiQue to evaluate the performance gains introduced by our methodology. 

Given the high computational expense associated with LLM-based knowledge graph construction, we restrict our evaluation to a random subset of 500 questions per benchmark.

\subsection{Baselines}
In the experiments, we include three different types of baseline RAG methods against which we compare our proposed pipeline.

First, we employ the Naive RAG baseline, which retrieves the top-$k$ documents with the highest cosine similarity between their embeddings and the query embedding, and feeds them directly into the LLM’s prompt together with the input query. We evaluate three configurations, setting $k$ to 10, 20, and 30. The two LLMs used in our experiments have context windows of approximately 16k and 38k tokens (around 12,000 and 98,000 words). Meanwhile, Wikipedia passages in MuSiQue range from 20 to 300 words, and those in 2WikiMultihopQA are of similar length. This allows us to compare advanced methods with simple baselines that leverage the extended context windows of modern LLMs, and to examine how increasing the number of retrieved paragraphs affects the performance of the RAG system.

Second, we consider a system based on query decomposition. The LLM is prompted to break a complex question into a sequence of simpler, preferably single-hop subquestions of arbitrary length. The system then retrieves context and derives intermediate answers sequentially. For the first subquestion, it retrieves $k$ passages to answer it. For each subsequent subquestion, it retrieves passages in the same way and generates an answer using the previous subquestions and their answers as additional context. After processing the full sequence, the system combines the intermediate questions and answers with the top-$k$ documents retrieved for the original question and uses this context to derive the final answer. We set $k=10$ in this baseline.

Finally, we use \textit{HippoRAG 2} as a graph-based baseline, specifically the second version proposed in~\cite{gutierrez2025rag}, for which state-of-the-art performance is reported on the MHQA benchmarks used in our evaluation. We use the default parameter configuration, which was tuned on a sample of 100 questions from the MuSiQue training set (see Appendix G of~\cite{gutierrez2025rag} for details).

\section{Results}
In this section, we evaluate SA-RAG on the previously described benchmarks and compare it against the baseline methods introduced in Section 4.2. We report Exact Match (EM) together with F1, an overlap-based metric that awards partial credit when the predicted answer shares content with the reference. To evaluate semantic accuracy beyond lexical overlap, we also report G-Eval Correctness~\cite{liu2023geval}, which uses an LLM judge with chain-of-thought reasoning to score answers against the ground truth on a continuous scale from $0.0$ to $1.0$; this metric tends to provide more accurate scores in cases where the model produced semantically correct responses (e.g., valid aliases) that EM failed to reward, as well as for answers that obtained high F1 scores due to partial lexical overlap while remaining factually incorrect. 

Recent studies have shown that, when LLMs with long context windows are used, a brute-force approach that provides the model with a large portion of the document corpus may outperform dense retrieval over a limited number of documents~\cite{fan2026we}. In addition, using larger LLM backbones may narrow the performance gap between graph-based methods and those that rely on dense vector retrieval~\cite{li2024retrieval}. Therefore, we further analyze input token consumption in Subsection~\ref{sub:token}. 

Additional ablation studies on token counts and component removal are reported in Appendix~\ref{app:ablation}

\subsection{Main results}

\begin{table}[t]
\centering
\caption{Evaluation results across (a) MuSiQue and (b) 2WikiMultiHopQA.}
\label{tab:rag_evaluation}
\normalsize
\setlength{\tabcolsep}{2pt}
\renewcommand{\arraystretch}{1.0}

\textbf{(a) MuSiQue}

\begin{tabular}{@{}llccc@{}}
\toprule
\textbf{Model} & \textbf{Methodology} & \textbf{EM} &
\textbf{F1} & \textbf{Correctness} \\
\midrule
phi4  & Naive RAG ($k=10$)  & 0.186 & 0.295 & 0.334 \\
      & Naive RAG ($k=20$)  & 0.208 & 0.326 & 0.412 \\
      & Naive RAG ($k=30$)  & \underline{0.228} & \textbf{0.355} & \underline{0.442} \\
      & Query-decomposition & 0.224 & 0.332 & 0.405 \\
      & HippoRAG 2          & 0.186 & 0.339 & 0.421 \\
      & SA-RAG              & \textbf{0.232} & \underline{0.348} & \textbf{0.449} \\
\midrule
gemma3 & Naive RAG ($k=10$)  & 0.21 & 0.301 & 0.384 \\
      & Naive RAG ($k=20$)   & 0.24  & 0.343  & 0.429 \\
      & Naive RAG ($k=30$)   & \textbf{0.278} & \textbf{0.381} & \textbf{0.454} \\
       & Query-decomposition & 0.242 & 0.353 & 0.436 \\
       & HippoRAG 2          & 0.21 & 0.317 & 0.390 \\
       & SA-RAG              & \underline{0.258} & \underline{0.362} & \underline{0.443} \\
\bottomrule
\end{tabular}

\vspace{0.8em}

\textbf{(b) 2WikiMultiHopQA}

\begin{tabular}{@{}llccc@{}}
\toprule
\textbf{Model} & \textbf{Methodology} & \textbf{EM} &
\textbf{F1} & \textbf{Correctness} \\
\midrule
phi4  & Naive RAG ($k=10$)  & 0.416 & 0.484 & 0.507 \\
      & Naive RAG ($k=20$)  & 0.44 & 0.513 & 0.549 \\
      & Naive RAG ($k=30$)  & \textbf{0.418} & 0.502 & 0.545 \\
      & Query-decomposition  & 0.372 & 0.432 & 0.479 \\
      & HippoRAG 2          & \underline{0.404} & \textbf{0.545} & \textbf{0.659} \\
      & SA-RAG               & 0.402 & \underline{0.513} & \underline{0.584}\\
\midrule
gemma3 & Naive RAG ($k=10$)  & 0.42 & 0.487 & 0.549 \\  
       & Naive RAG ($k=20$)   & 0.444 & 0.515 & 0.581 \\
       & Naive RAG ($k=30$)  & 0.438 & 0.513 & 0.584 \\
       & Query-decomposition & 0.398 & 0.457 & 0.517 \\
       & HippoRAG 2          & \textbf{0.484} & \underline{0.571} & \underline{0.615} \\
       & SA-RAG              & \underline{0.478} & \textbf{0.574} & \textbf{0.649} \\
\bottomrule
\end{tabular}

\end{table}

Table~\ref{tab:rag_evaluation} presents the experimental results, comparing the performance of different RAG approaches across the benchmarks. The experiments were conducted using two LLMs for reasoning and answer generation, \textit{phi4} and \textit{gemma3}, and \textit{qwen3-embedding:0.6b} was used to generate the text embeddings. For each setup, the same LLM used to answer the questions was also used to construct the knowledge graph during indexing.

The first observation is that SA-RAG ranks among the two best-performing methods across all settings, according to both F1 and correctness. On MuSiQue, it outperforms all baselines when \textit{phi4} is used; with \textit{gemma3}, it ranks second to vector-based RAG with $k=30$. On 2WikiMultihopQA, SA-RAG performs best with \textit{gemma3} and ranks second to HippoRAG 2 with \textit{phi4}. 

Second, except on 2WikiMultiHopQA with \textit{phi4}, SA-RAG outperforms HippoRAG 2 in every setting. This performance gap is more pronounced on MuSiQue and is consistent across all metrics. However, G-Eval is inherently stochastic, so the improvements in correctness scores do not provide definitive evidence of better performance. Nevertheless, the results support the conclusion that SA-RAG achieves competitive performance compared with a state-of-the-art GraphRAG method.

Table~\ref{tab:rag_evaluation} presents the experimental results, comparing the performance of different RAG approaches across the benchmarks. We conducted the experiments using two LLMs, \textit{phi4} and \textit{gemma3}, for reasoning and answer generation, and used \textit{qwen3-embedding:0.6b} to generate text embeddings. For each setup, the same LLM used to answer the questions was also used to construct the knowledge graph during indexing.

The first observation is that SA-RAG ranks among the two best-performing methods across all settings, according to both F1 and correctness. On MuSiQue, it outperforms all baselines with \textit{phi4}; with \textit{gemma3}, it ranks second to vector-based RAG with $k=30$. On 2WikiMultiHopQA, SA-RAG performs best with \textit{gemma3} and ranks second to HippoRAG 2 with \textit{phi4}.

Second, except on 2WikiMultiHopQA with \textit{phi4}, SA-RAG outperforms HippoRAG 2 in every setting. This performance gap is more pronounced on MuSiQue and is consistent across all metrics. However, because G-Eval is inherently stochastic, the improvements in correctness scores do not provide definitive evidence of better performance. Nevertheless, the results support the conclusion that SA-RAG performs competitively with a state-of-the-art GraphRAG method.

An interesting and somewhat surprising result is that increasing the number of retrieved document passages, $k$, yields substantial performance gains on MuSiQue. With \textit{gemma3}, this approach even outperforms all the baselines. By contrast, on 2WikiMultiHopQA, the performance gains from increasing $k$ diminish. We also observe that query decomposition performs worse on 2WikiMultiHopQA than the simplest baseline. One possible explanation lies in how the benchmarks were constructed. Although MuSiQue is generally more challenging, with stronger dependencies between reasoning steps and stricter guarantees that its questions are genuinely multi-hop, retrieval coverage appears to be a major challenge for RAG systems on this benchmark. Advanced methods may improve retrieval efficiency, but sufficiently broad dense retrieval can compensate for their limitations. In contrast, 2WikiMultiHopQA comprises different question types that require different reasoning operations. For example, inference questions require deriving new information from retrieved facts, whereas bridge-comparison questions require comparing properties of intermediate entities. Thus, the main challenge may not be reasoning or evidence integration; increasing the context size may also introduce more noise into the inference process.

\subsection{Context Size and Token Efficiency}
\label{sub:token}

To investigate whether the performance gains result solely from retrieving more context, we examine the number of input tokens per question across the benchmarks. For this analysis, we use the \textit{phi4} tokenizer; similar results are obtained with the \textit{gemma3} tokenizer (see Appendix~\ref{app:addfig}). Summary statistics are presented in Table~\ref{tab:tokens} (see Appendix~\ref{app:addfig} for additional visualizations).

\begin{table}[htbp]
  \centering
  \caption{Token usage across (a) MuSiQue and (b) 2WikiMultiHopQA. Capped total, mean, and standard deviation are calculated after limiting each input context to fit within \textit{phi4}’s context window.}
  \label{tab:tokens}

  \begin{tabular}{lcccc}
    \multicolumn{5}{c}{(a) MuSiQuE} \\[0.4em]
    \toprule
    method & total & total capped & mean capped & SD capped \\
    \midrule
    Naive (20) & 1,281,889 & 1,281,889 & 2,563.8 & 543.9 \\
    SA-RAG & 1,799,352 & 1,652,889 & 3,305.8 & 4,083.4 \\
    Naive (30) & 1,919,116 & 1,919,116 & 3,838.2 & 729.4 \\
    \bottomrule
  \end{tabular}

  \vspace{1.2em}

  \begin{tabular}{lcccc}
    \multicolumn{5}{c}{(b) 2WikiMultiHopQA} \\[0.4em]
    \toprule
    method & total & total capped & mean capped & SD capped \\
    \midrule
    Naive (20) & 1,412,166 & 1,412,166 & 2,824.3 & 699.5 \\
    SA-RAG & 2,194,746 & 1,982,975 & 3,966.0 & 4,736.8 \\
    Naive (30) & 2,155,891 & 2,155,891 & 4,311.8 & 912.5 \\
    \bottomrule
  \end{tabular}
\end{table}

As shown in the table, on MuSiQue, our system retrieves more tokens than vector-based retrieval with $k=20$ but fewer than with $k=30$, both in total and on average. On 2WikiMultiHopQA, the overall number of retrieved tokens is the highest among the methods considered. However, when we count only the tokens that fit within \textit{phi4}’s context window, the total and average context sizes again fall between those of the vector-based retrievers with $k=20$ and $k=30$.

The most pronounced differences in token utilization between our method and vector-based methods are reflected in the standard deviations, indicating substantially greater variation in our method’s token use across questions. On MuSiQue, our method’s standard deviation is 4,083 tokens, roughly five to eight times the vector-based methods' values of 543 and 779; on 2WikiMultiHopQA, it is 4,713 tokens, again roughly five to seven times the vector-based methods' values of 669 and 912.

\begin{figure}[htbp]
    \centering
    \includegraphics[width=\columnwidth]{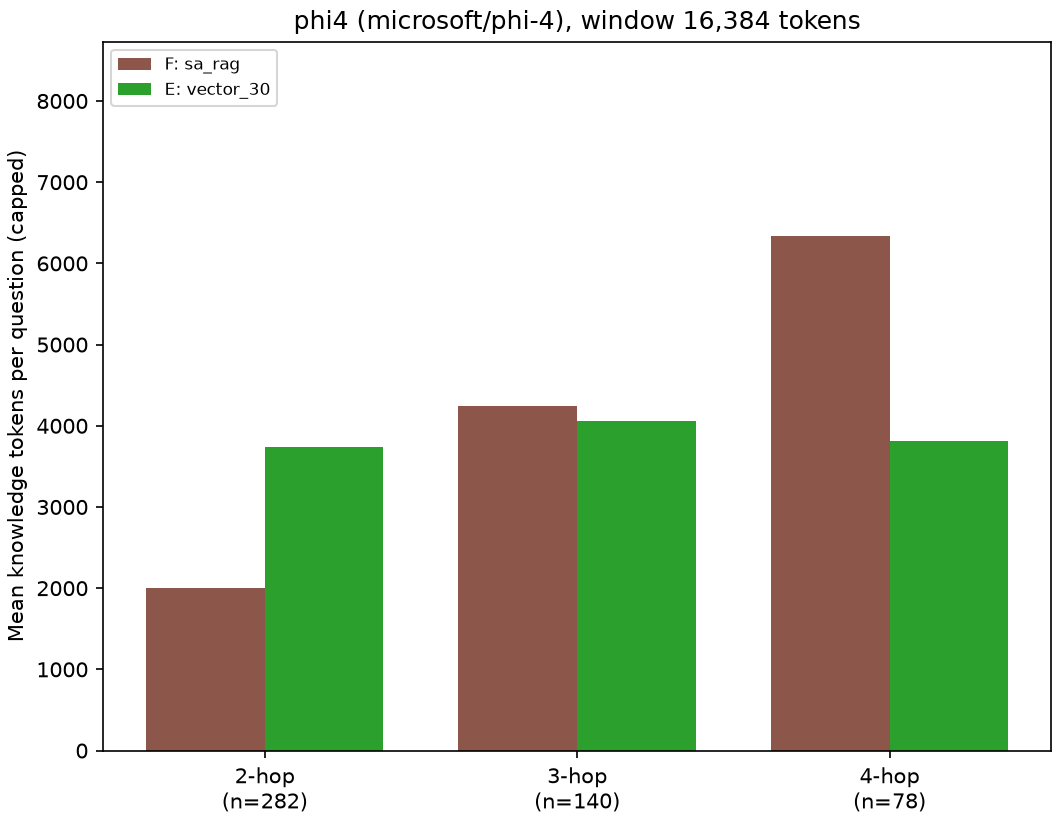}
    \caption{Mean tokens in input context per question category in MuSiQue}
    \label{fig:hop_tokens}
\end{figure}

Since MuSiQue includes questions requiring different numbers of reasoning steps, and the number of steps is encoded in each question's ID, we also analyze the average number of input tokens for two, three, and four-hop questions in our test sample. We compare SA-RAG with Naive RAG using $k=30$.  The results are shown in Figure~\ref{fig:hop_tokens}. For SA-RAG, the average number of input tokens increases with the number of hops: 2,003 for two-hop questions, 4,238 for three-hop questions, and 6,341 for four-hop questions. In contrast, dense vector retrieval uses a similar number of tokens across hop categories, ranging from 3,733 to 4,064. This suggests that the retrieval volume of SA-RAG scales with question complexity, whereas that of Naive RAG remains roughly constant regardless of hop count.

\section{Conclusions}
In this paper, we addressed the challenges of document retrieval in RAG systems for complex tasks that require multistep reasoning and the aggregation of evidence from multiple documents. We proposed integrating the SA algorithm into RAG approaches based on heterogeneous knowledge graphs to address the limitations of current systems without expensive language model fine-tuning. Although the algorithm has traditionally been used in other information retrieval settings alongside sparse or dense vector retrievers, we investigated whether it could serve as a useful component of graph-based RAG systems. Our experiments demonstrate the potential of this approach, with performance equivalent to or better than that of a competitive GraphRAG baseline and dense vector retrieval with a large context window.

However, several limitations of the current work should be acknowledged and addressed in future research. First, due to budget constraints and high experimental costs, we compared our method against only a few baselines. More extensive validation would help establish the robustness and generalizability of our findings. Second, when assigning association weights to relationship links in the knowledge graph, we used cosine similarity values from an off-the-shelf embedding model and adapted them only through simple linear scaling for our task. Future work should explore improvements to the retrieval components, such as fine-tuning the embedding model for better task alignment. Third, future work could investigate soft prompting approaches that use retrieved information to guide answer generation. For example, a GNN could encode the retrieved subgraph to generate soft prompts for the LLM, while an RNN could similarly process a sequence of retrieved relationship triplets.

\appendices
\section*{ Acknoweledgemnt}
Miklós Krész was supported by the Slovenian Research and Innovation Agency (ARIS) through research program P1-0404 and by the research program Cognicom (0013103) at the University of Primorska. He is also grateful for the support of ARIS grants J7-60122 and BI-HU/26-27-006.

\section{Motivational example}

\label{app:motivational}
Equation~\ref{eq:centrality} shows the exact formula used to calculate the closeness centrality scores for each node $v$. Similarly, Equation~\ref{eq:ppr} shows the formula for Personalized PageRank using the same notation. As discussed in Section~\ref{sub:motiv}, both sets of scores were subsequently scaled to the $[0, 1]$ interval using min--max normalization.

\begin{equation}\label{eq:centrality}
\begin{aligned}
C(v) &=
\begin{cases}
|S_v|/D_v, & D_v > 0,\\
0,          & D_v = 0,
\end{cases}\\
D_v &= \sum_{s \in S_v} d(v,s),\\
d(v,s) &= \min_{P:v\leadsto s}
          \sum_{(u,w)\in P} d_{u,w},\\
d_{u,w} &= 1 - s_{uw}.
\end{aligned}
\end{equation}

\begin{equation}\label{eq:ppr}
\begin{aligned}
\mathrm{PPR}(v) &=
\begin{cases}
\begin{aligned}
&(1-\alpha)\mathbb{I}(v \in S_v) \\
&\quad + \alpha \sum_{u \in \mathcal{N}(v)} w_{uv}\mathrm{PPR}(u),
\end{aligned} & |S_v| > 0,\\[8pt]
0, & |S_v| = 0,
\end{cases}\\[4pt]
w_{uv} &= s_{uv}\Big/\sum_{w \in \mathcal{N}(u)} s_{uw}.
\end{aligned}
\end{equation}

Here, \(S_v\) is the set of reachable seed nodes, and \(s_{uw}\) is the similarity score of the edge between \(u\) and \(w\). In~\eqref{eq:centrality}, \(d_{u,w}\) denotes the distance of edge \((u,w)\), \(d(v,s)\) is the shortest-path distance from \(v\) to \(s\) over paths \(P: v \leadsto s\), \(D_v\) is the total distance from \(v\) to all reachable seed nodes in \(S_v\), and \(C(v)\) is the resulting closeness centrality score (set to \(0\) when \(D_v = 0\)). In~\eqref{eq:ppr}, \(\mathcal{N}(u)\) denotes the neighborhood of node \(u\), \(w_{uv}\) is the normalized edge weight between \(u\) and \(v\), \(\alpha\) is the damping factor, \(\mathbb{I}(\cdot)\) is the indicator function, and \(\mathrm{PPR}(v)\) is the Personalized PageRank score of node \(v\) (set to \(0\) when \(|S_v| = 0\)).

\section{Effects of edge weight normalization}
\label{app:normalization}

\begin{figure}[htbp]
  \centering
  \includegraphics[width=\columnwidth]{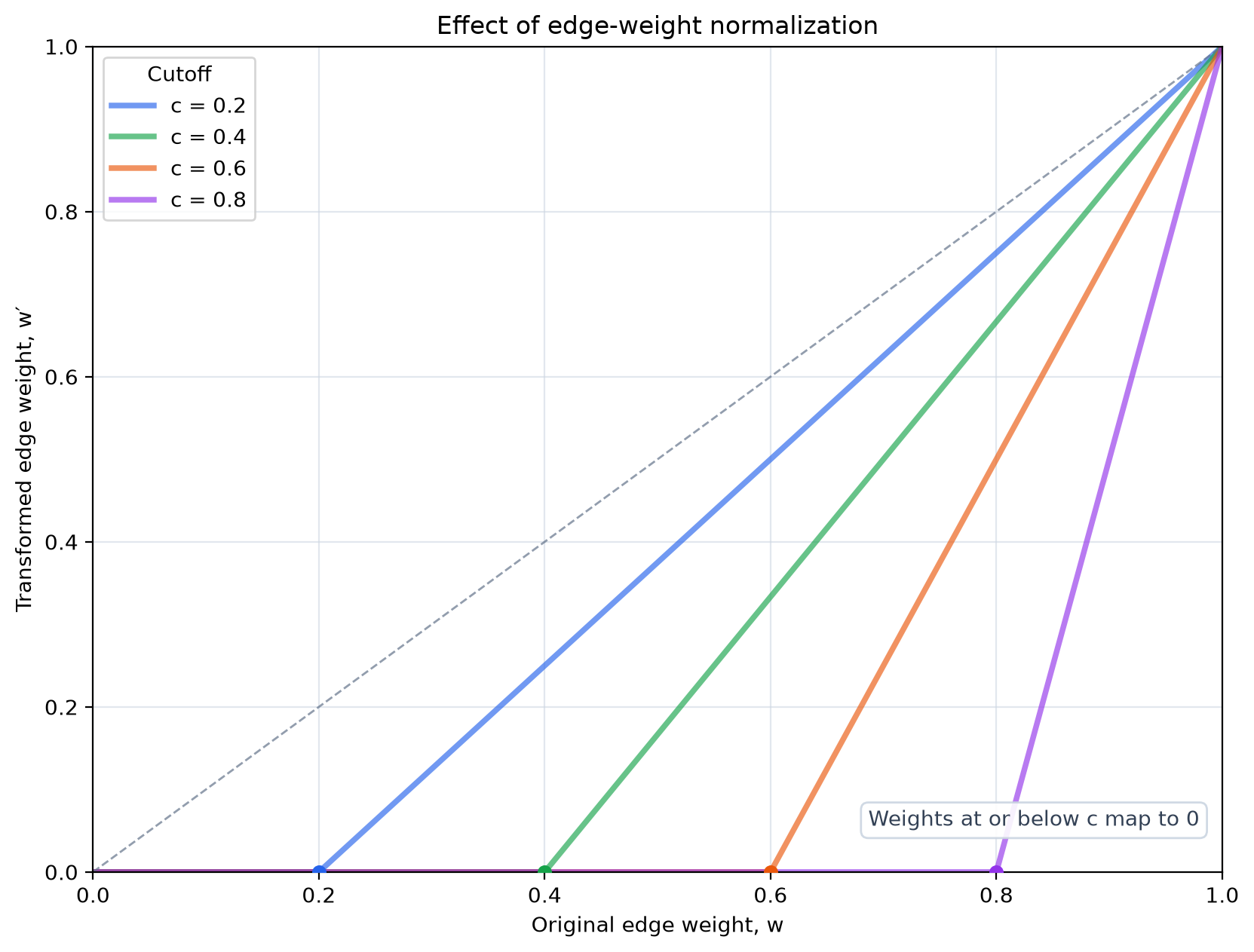}
  \caption{Piecewise-linear scaling functions for edge weights across different values of the normalization constant $c$.}
  \label{fig:sanorm1}
\end{figure}

\begin{figure}[htbp]
  \centering
  \includegraphics[width=\columnwidth]{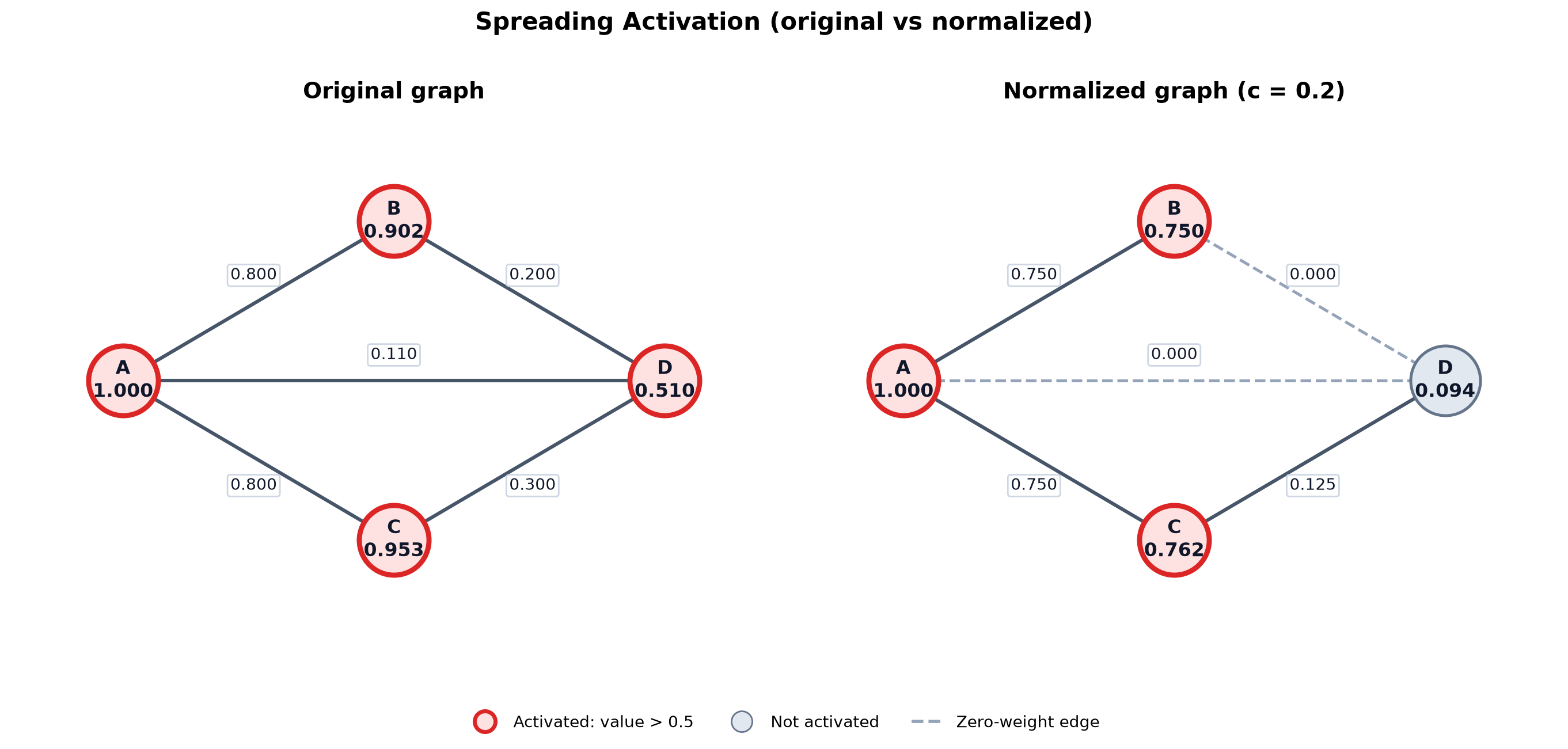}
  \caption{Outcomes of the spreading activation process starting from node $A$ on the original weighted graph (left) and the graph with rescaled edge weights ($c = 0.2$, right); activated nodes are highlighted in red.}
  \label{fig:sanorm2}
\end{figure}

This section illustrates the effect of edge-weight normalization used in our system to prevent context explosion in spreading activation; Figure~\ref{fig:sanorm1} plots the rescaling functions $y = \max\!\left(0, \frac{x - c}{1 - c}\right)$ for $x \in [0, 1]$ across different values of the normalization constant $c$, and Figure~\ref{fig:sanorm2} compares the resulting node activations on a four-node weighted graph before and after rescaling edge weights with $c = 0.2$.

\section{Additional Visualizations: Token Usage}
\label{app:addfig}

\begin{figure}[htbp]
  \centering
  \includegraphics[width=\columnwidth]{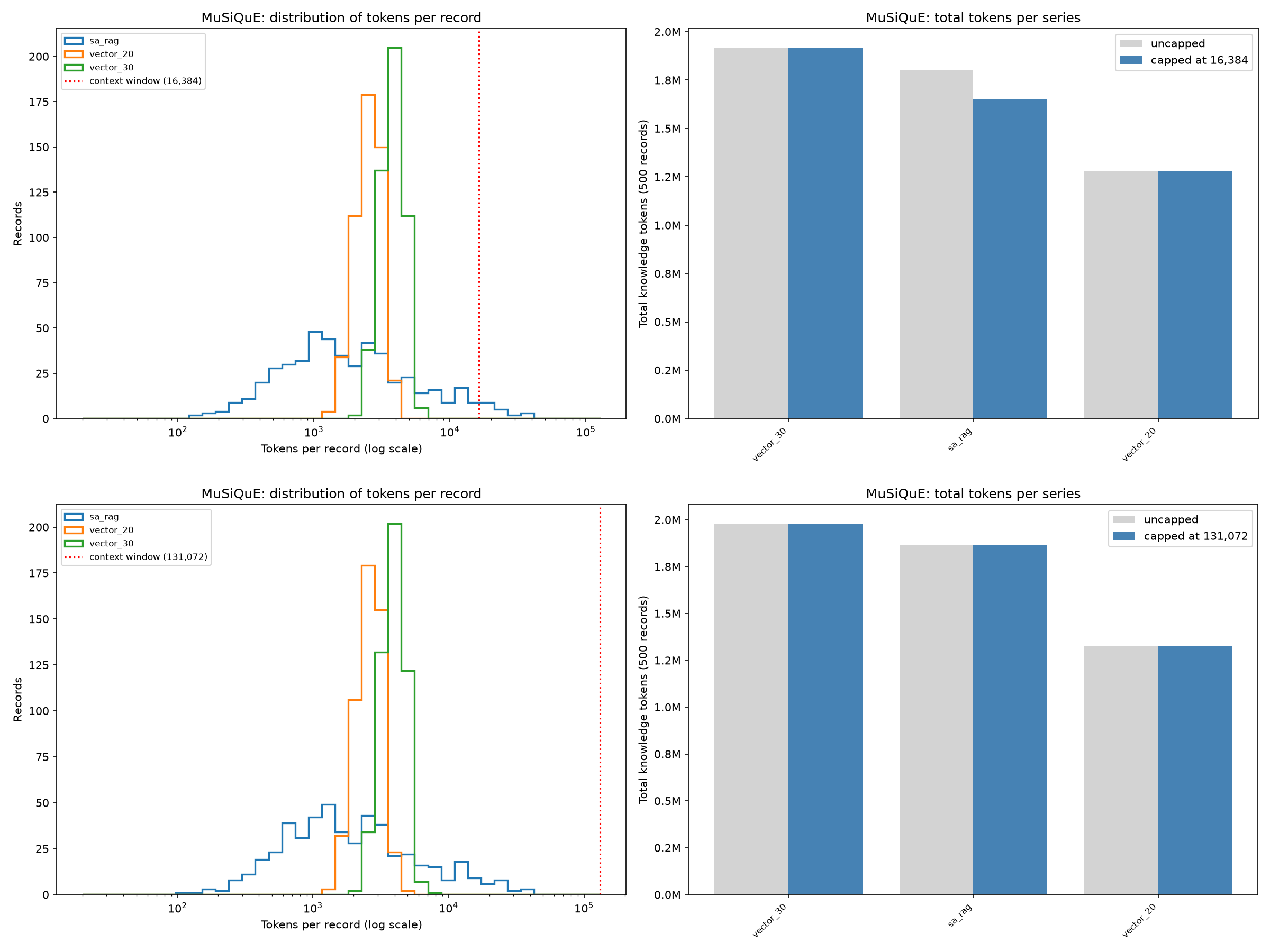}
  \caption{Distribution of knowledge tokens alongside total capped and uncapped tokens for MuSiQue questions, evaluated using the tokenizers of \textit{phi4} (top) and \textit{gemma3} (bottom).}
  \label{fig:tkndst}
\end{figure}

\begin{figure}[htbp]
  \centering
  \includegraphics[width=\columnwidth]{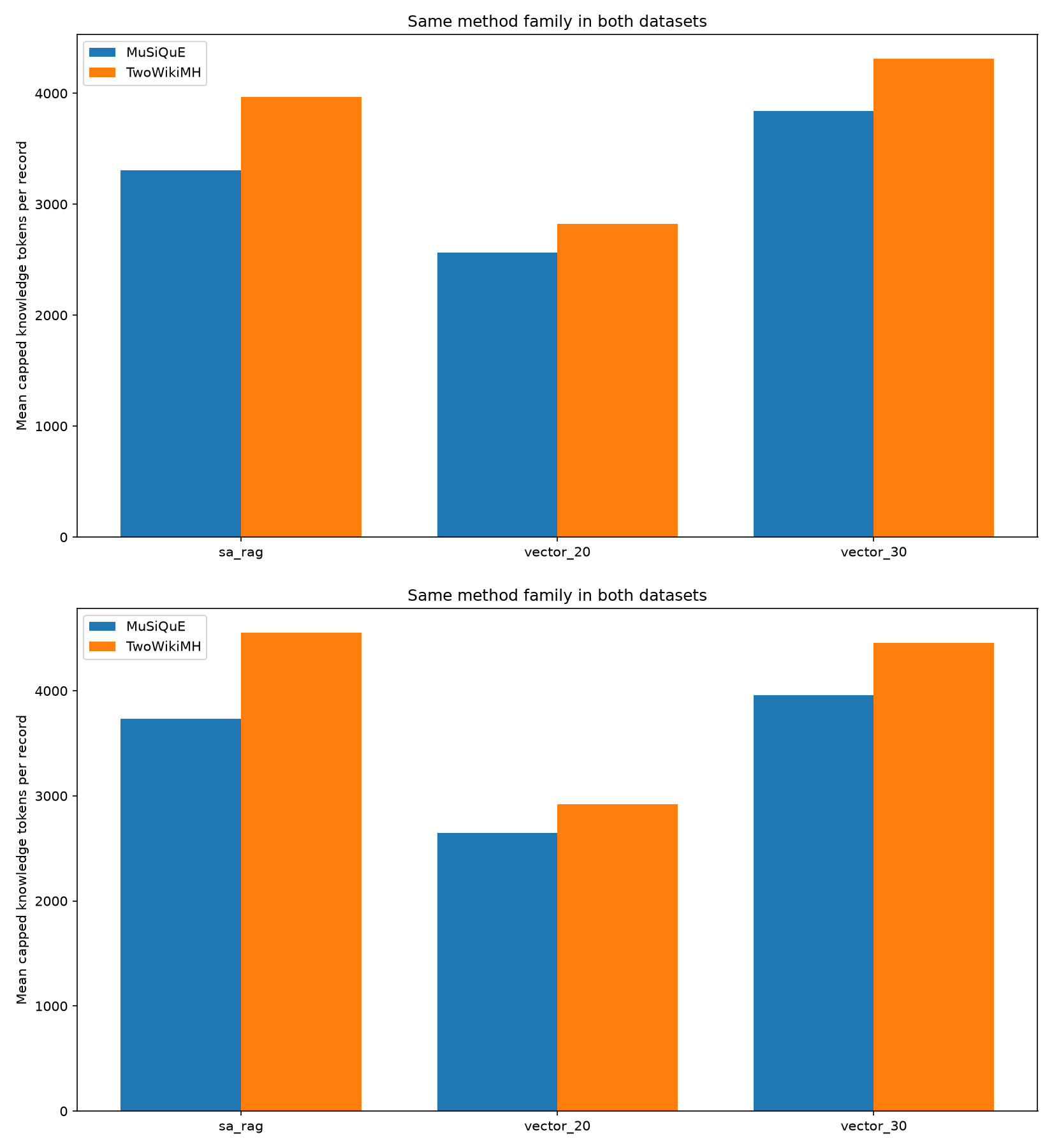}
  \caption{Average number of capped knowledge tokens used on the MuSiQue and 2WikiMultiHopQA benchmarks for \textit{phi4} (top) and \textit{gemma3} (bottom).}
  \label{fig:tknusg}
\end{figure}

Figure~\ref{fig:tkndst} displays the distribution of knowledge tokens as well as total capped and uncapped tokens for MuSiQue questions across the tokenizers of the \textit{phi4} and \textit{gemma3} models. Figure~\ref{fig:tknusg} shows the average (capped) number of knowledge tokens used by \textit{gemma3} and \textit{phi4} on the MuSiQue and 2WikiMultiHopQA benchmarks.

\section{Ablation study}
\label{app:ablation}
This section reports supplementary ablation results on the MuSiQue and 2WikiMultiHopQA benchmarks. Table~\ref{tab:ablation_tokens} summarizes the effect of removing the SA-based pruning stage on total and mean capped input token counts (with \textit{phi4} tokenizer). Table~\ref{tab:ablation_rag_evaluation} presents the corresponding Exact Match (EM), F1, and Correctness scores for \textit{phi4} and \textit{gemma3} when removing either SA-based pruning or relationship context

\begin{table}[htbp]
  \centering
  \caption{Effect of removing the SA-based pruning stage on average and total input token usage across the MuSiQue and 2WikiMultiHopQA benchmarks.}
  \label{tab:ablation_tokens}

  \begin{tabular}{lcc}
    \multicolumn{3}{c}{(a) MuSiQuE} \\[0.4em]
    \toprule
    method & total capped & mean capped\\
    \midrule
    Original & 1,652,889 &  3,305.8 \\
    w/o SA-based pruning & 4,762,002 & 9,524.0 \\
    \bottomrule
  \end{tabular}

  \vspace{1.2em}

  \begin{tabular}{lcc}
    \multicolumn{3}{c}{(b) 2WikiMultiHopQA} \\[0.4em]
    \toprule
    method & total capped & mean capped \\
    \midrule
    Original & 1,982,975 & 3,965.9 \\
    w/o SA-based pruning & 3,699,944 & 7,399.9 \\
    \bottomrule
  \end{tabular}
\end{table}

\begin{table}[htbp]
  \centering
  \caption{Ablation results on MuSiQue and 2WikiMultiHopQA evaluating the removal of SA-based graph pruning and relationship context.}
 \label{tab:ablation_rag_evaluation}
 \normalsize
 \setlength{\tabcolsep}{2pt}
 \renewcommand{\arraystretch}{1.0}
 
 \textbf{(a) MuSiQue}
 
 \begin{tabular}{@{}llccc@{}}
 \toprule
 \textbf{Model} & \textbf{Methodology} & \textbf{EM} &
 \textbf{F1} & \textbf{Correctness} \\
 \midrule
 phi4  & Original & 0.232 & 0.348 & 0.449 \\
      & w/o SA-based pruning & 0.158 & 0.253 & 0.354 \\
      & w/o Relationship context & 0.214 & 0.338 & 0.433 \\
 \midrule
 gemma3 & Original & 0.258 & 0.362 & 0.443 \\
      &  w/o SA-based pruning & 0.262 & 0.350 & 0.429 \\
      & w/o Relationship context & 0.246 & 0.353 & 0.433 \\   
 \bottomrule
 \end{tabular}
 \vspace{0.8em}
 
 \textbf{(b) 2WikiMultiHopQA}
 \begin{tabular}{@{}llccc@{}}
 \toprule
 \textbf{Model} & \textbf{Methodology} & \textbf{EM} &
 \textbf{F1} & \textbf{Correctness} \\
 \midrule
 phi4  & Original & 0.402 & 0.513 & 0.584 \\
      & w/o SA-based pruning & 0.378 & 0.481 & 0.538 \\
      & w/o Relationship context & 0.416 & 0.523 & 0.58 \\  
 \midrule
 gemma3 & Original & 0.478 & 0.574 & 0.649 \\
      & w/o SA-based pruning & 0.466 & 0.557 & 0.623 \\
      & w/o Relationship context & 0.416 & 0.520 & 0.598 \\
 \bottomrule
 \end{tabular}
 \end{table}

As we can see from table~\ref{tab:ablation_tokens}, removing the SA-based pruning stage substantially increases capped input token counts across both datasets. Specifically, total and mean input tokens roughly triple on MuSiQue (from 1,652,889 to 4,762,002 total) and nearly double on 2WikiMultiHopQA (from 1,982,975 to 3,699,944 total).

Table~\ref{tab:ablation_rag_evaluation} shows that removing either SA-based pruning or relationship triplets from the knowledge context decreases Correctness and F1 scores across almost all settings. The only exception is \textit{phi4} on 2WikiMultiHopQA, where only the removal of the pruning stage degrades performance. This exception likely occurs because a large normalization constant ($c=0.7$) retrieves fewer relationships compared to MuSiQue; combined with \textit{phi4}'s limited context window and a large volume of retrieved documents, most relationship triplets are truncated from the context anyway (hough verifying this hypothesis requires further investigation). Howerver in all the other cases, decreases in performance are notibale, suggesting that both components bring added value to the system (this is especially true for SA-based pruning when the system is used with \textit{phi4} or other models with smaller context windows).

\clearpage
\bibliographystyle{unsrt}
\bibliography{references}
\end{document}